\documentclass[letterpaper, 10 pt, conference]{ieeeconf}  
\IEEEoverridecommandlockouts                              
\usepackage{gensymb}
\usepackage{cclicenses, graphicx}
\usepackage{amsmath}
\usepackage{array}
\usepackage{flushend}
\usepackage{flushend}

\newcolumntype{C}[1]{>{\centering\arraybackslash}m{#1}}

\title{\LARGE \bf Design and Development of Effective Transmission Mechanisms\\ on a Tendon Driven Hand Orthosis for Stroke Patients}

\author{Sangwoo Park$^{1}$, Lynne Weber$^{2}$, Lauri Bishop$^{2}$, Joel Stein$^{2,3}$ and Matei Ciocarlie$^{1,3}$%
\thanks{*This work was supported in part by the Columbia-Coulter Translational
Research Partnership and the National Science Foundation under grant IIS-1526960 (part of the National Robotics Initiative).}%
\thanks{$^{1}$Department of Mechanical Engineering, Columbia University, New York, NY 10027, USA.}%
\thanks{\hspace{-3mm}{\tt\small \{sp3287, matei.ciocarlie\}@columbia.edu}}%
\thanks{$^{2}$Department of Rehabilitation and Regenerative Medicine, Columbia University, New York, NY 10027, USA. {\tt\small \{lw2739, lb2413, js1165\}@cumc.columbia.edu}}%
\thanks{$^{3}$Co-Principal Investigators}
}

\begin{document}

\maketitle
\thispagestyle{empty}
\pagestyle{empty}

\begin{abstract}
Tendon-driven hand orthoses have advantages over exoskeletons with respect to wearability and safety because of their low-profile design and ability to fit a range of patients without requiring custom joint alignment. However, no existing study on a wearable tendon-driven hand orthosis for stroke patients presents evidence that such devices can overcome spasticity given repeated use and fatigue, or discusses transmission efficiency. In this study, we propose two designs that provide effective force transmission by increasing moment arms around finger joints. We evaluate the designs with geometric models and experiment using a 3D-printed artificial finger to find force and joint angle characteristics of the suggested structures. We also perform clinical tests with stroke patients to demonstrate the feasibility of the designs. The testing supports the hypothesis that the proposed designs efficiently elicit extension of the digits in patients with spasticity as compared to existing baselines.

\end{abstract}

\section{Introduction}
One of the most common impairment patterns in stroke patients is loss of fine motor control together with spasticity that limits functional use of the hand~\cite{cauraugh2000}. In recent years, many wearable robots have been developed for patients with incomplete motor recovery, a condition which affects more than half of stroke patients following conventional rehabilitation~\cite{schaechter2004}. In designing these wearable robots, the focus has often been on providing the patients a large number of exercise repetitions. In order to deliver sufficient assistance so that patients are able to extend their hand against the force of spasticity, the actuation also has to be strong enough while keeping the device compact.

Linkage-based exoskeletons generally provide efficient force transmission; however, alignment of an axis of rotation with each finger and its respective robotic joint remains challenging~\cite{stienen2009}. Misalignment can result in discomfort and even injury~\cite{schiele2006}.

In contrast, a tendon-driven system does not require alignment of the joints. These systems are also better suited for underactuation, allowing a design with a small number of motors to reduce the overall device size. However, because of inefficient transmission compared with exoskeletons, larger and more powerful motors may be needed to overcome spasticity.

\begin{figure}[t]
\centering
\begin{tabular}{c}
\includegraphics[width=1.0\linewidth]{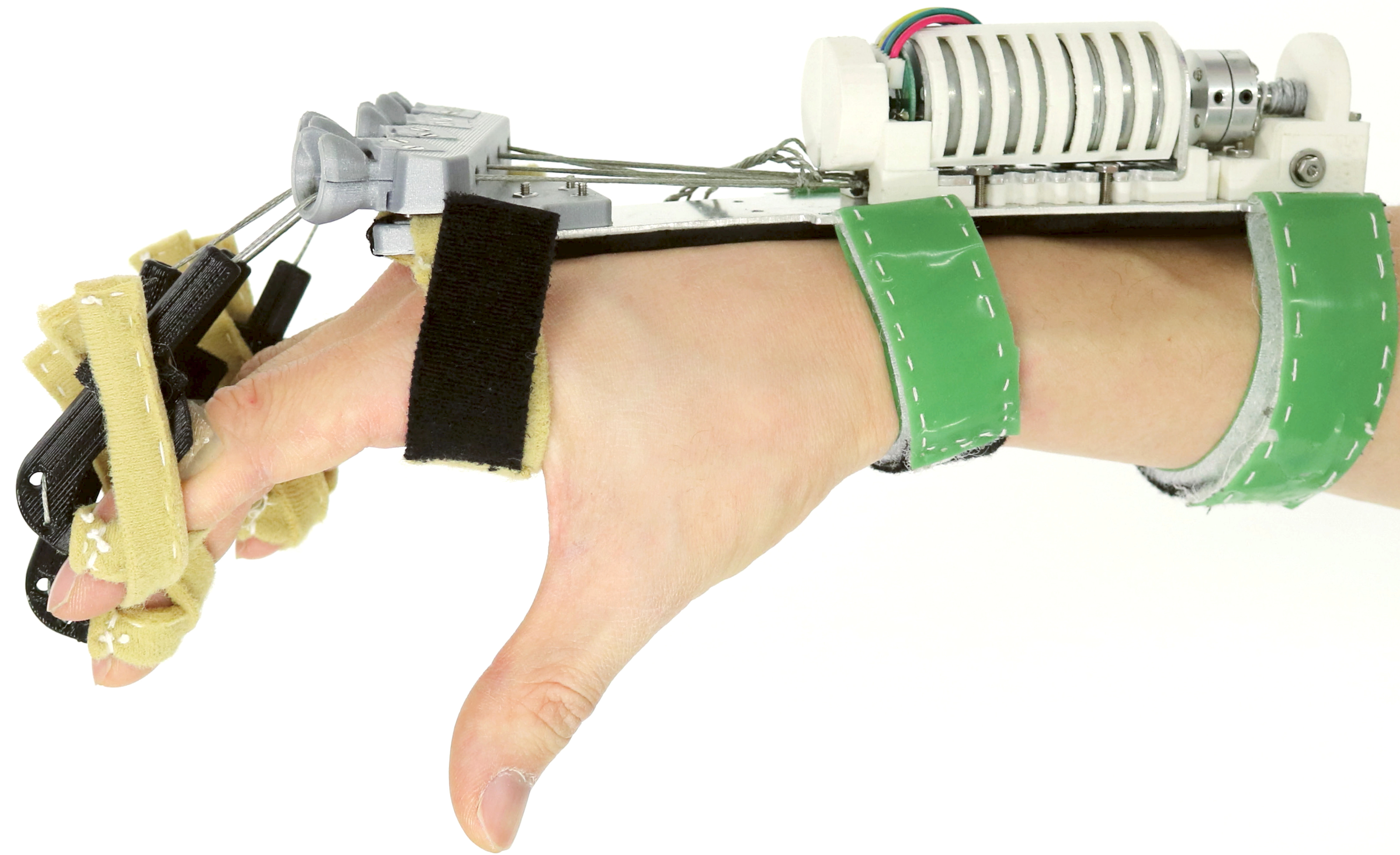}\\
\includegraphics[width=1.0\linewidth]{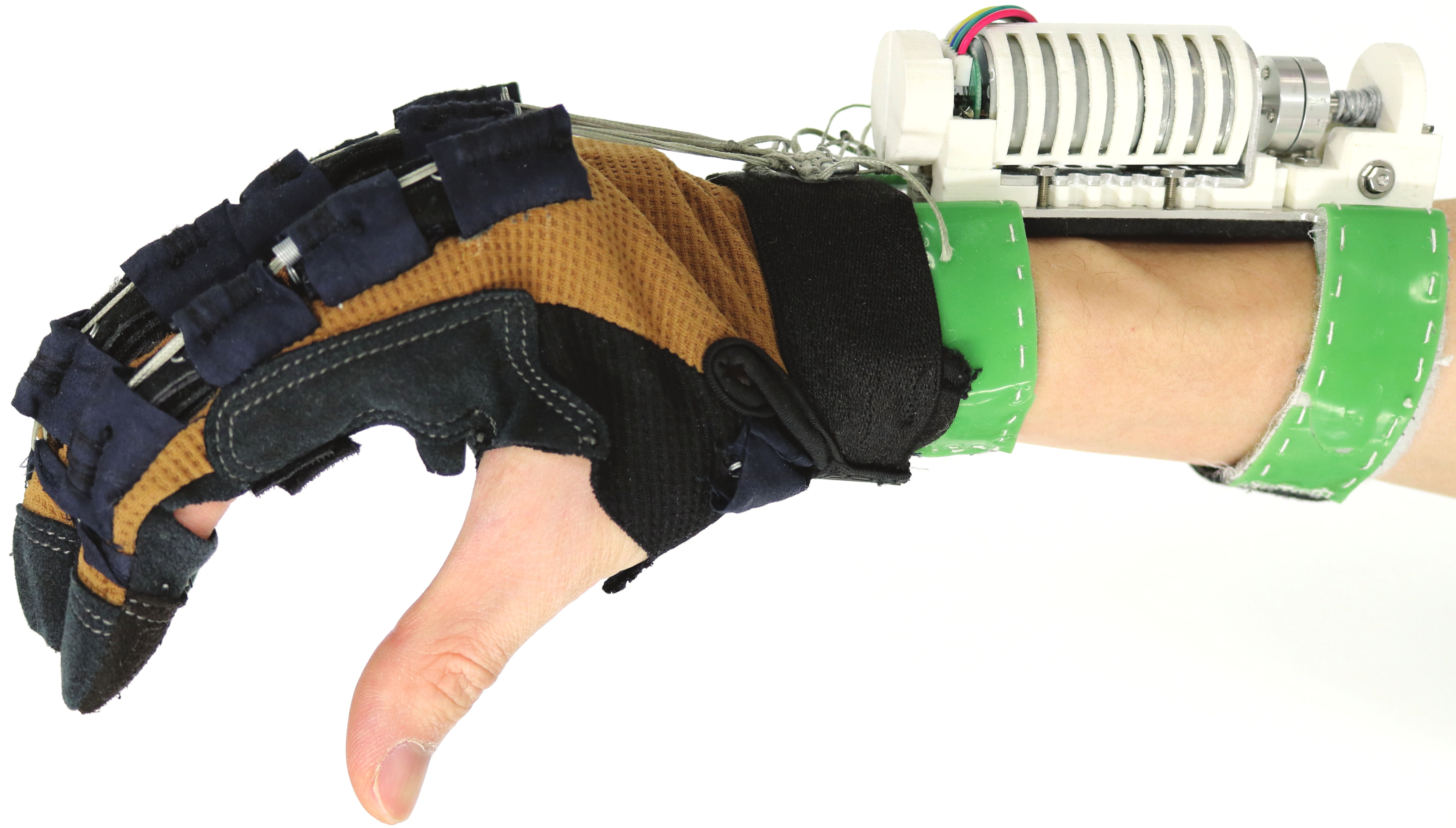}
\end{tabular}
\caption{Prototypes of the hand orthosis transmission mechanisms (top: Design A, bottom: Design B) proposed in this study.}
\label{fig:orthosis}
\end{figure}

In this study, we focus on the design of mechanical structures on a hand orthosis to enhance force transmission efficiency in a tendon-driven device. To achieve this, the device must assist hand opening with greater torques around each joint making the most of the given motor force without compromising wearability. Overall, the main contributions of this paper are the following:
\begin{itemize}
\item We propose two mechanical structure designs: one for higher spasticity at the proximal interphalangeal (PIP) joint than the metacarpophalangeal(MCP) joint, and one for equally severe tone on each joint.
\item Using mathematical models of the two designs, we measure the moment arms around the PIP and MCP joints. Assuming that the PIP and MCP joints move simultaneously, we vary design parameters to see how they affect the moment arms. Also, we compare between the two and a baseline design to demonstrate their validity.
\item We evaluate the computational results through experiments, quantitatively assessing the joint angles and force on the actuated tendon with a 3D-printed artificial finger, designed to mimic the finger of a patient after stroke.
\item We present quantitative results through clinical tests with stroke patients to demonstrate the theoretical computations and outcome of experiments with the artificial finger.
\end{itemize}

To the best of our knowledge, this study is the first to present and
evaluate \textit{transmission mechanisms by which exotendons can
  overcome hand spasticity for functional tasks with low motor forces
  and no rigid joints}. Efficient transmission is important as it
allows a reduction in the tendon force required to achieve functional
finger extension. Lower tendon forces allow the use of smaller motors;
they also reduce the unwanted phenomenon of distal migration, where, due to the
applied forces, the motor component of the devices slides on the
forearm towards the hand. Both of these characteristics can lead to
more wearable devices.

\section{Related Work}
Kamper et al.~\cite{kamper2006} posit that over-excitement of the flexor muscles and decreased activation of the extensor muscles during finger extension may be a result of an overactive stretch response, resulting in involuntary grasp. Finger extension is an essential component of functional grasping tasks and a number of wearable hand robots have been developed to facilitate this movement for enhanced performance of activities of daily living. One of the most successful off-the-shelf products, Saebo Flex, utilizes passive underactuation to provide spring-assisted hand opening. A study reported there were meaningful clinical improvements for the majority of participants for the Action Research Arm Test (ARAT) and Upper Limb Motricity Index after a 12 week rehabilitation program~\cite{stuck2014}. HandSOME~\cite{chen2017} and SPO~\cite{ates2017} are also passive hand devices developed for impairment compensation following stroke. While such passive devices achieve a compact and lightweight design, they inherently interfere with finger flexion due to the constantly applied spring force.

The mechanical interference with finger flexion present with mechanical finger extension aids can be avoided using active hand orthoses. Jo et al. presented a single degree of freedom (DOF) exoskeleton using linkage structures that follow fingertip trajectories found from experiments with a motion capture system~\cite{jo2017}. Pu et al. developed five digits actuated device, Exo-finger, based on hand kinematics using a linkage driven system~\cite{pu2016}. ExoK’ab utilizes mechanical transmission components, such as worm gears and sliders with two micro motors for four fingers and one motor for the thumb to assist independent hand movement~\cite{sandoval2016}. These exoskeletons are vulnerable to fit issues for hands of various sizes due to a well-known difficulty in joint axes alignment. To fill this need, Cempini et al. developed a hand exoskeleton with an autonomous joint axes aligning mechanism by letting passive DOFs absorb any misplacement of the joint center~\cite{cempini2015}.

Wearable hand orthoses composed of soft structures are low profile, and reduce the likelihood of injury, compared to exoskeletons. Soft pneumatic actuator based devices take advantage of natural compliance and flexibility for better interaction with the human hand, but remain tethered to external air pressure sources. A low-cost soft orthotic glove produced by Zhao et al.~\cite{zhao2016} contains integrated optical strain sensors. The optical bend sensors provide real time feedback on how each finger moves, which can be of great benefit. A customized inconsistent bending profile
with variable stiffness can also be implemented to enhance the usability of this type of device~\cite{yap2015}.

Tendon-driven devices have advantages over the two aforementioned mechanisms in terms of wearability since actuators can be remotely positioned and structures located on the hand only require a few small anchor points, which make the system more suitable for underactuation. Compact and lightweight, BiomHED, for example, exploits artificial exotendons to mimic human finger movement~\cite{lee2014}. Biggar and Yao built a tendon driven robotic glove with suction cups on an inner glove as a cable guide using vacuum pressure~\cite{biggar2016}. Exo-Glove utilizes exotendon drive system on the surface of a glove adopting a differential mechanism to actuate a multiple fingers with a single motor~\cite{in2015}. In an effort to reduce control inaccuracy caused by compliance in fabric-based gloves, Exo-Glove Poly was developed~\cite{kang2016}. This device allows better fit through various adaptable features rather than relying on the compliance of soft material.

The tendon driven devices described above deploy either multiple motors closely located to the transmission~\cite{lee2014},~\cite{biggar2016} or a distally mounted single strong motor connected to the end effector via a Bowden cable~\cite{in2015},~\cite{kang2016}. However, installation of many motors for the assistance of one movement pattern is redundant, and Bowden cables introduce unnecessary friction. Although there has been a study on transmission mechanisms that provide a natural joint extension sequence using a tendon-driven orthoses~\cite{kim2017}, no prior study has systematically examined effective transmission methods for tendon driven devices. Our approach seeks to achieve efficient transmission on a tendon-driven system while maintaining a compact and lightweight design for optimal wearability.

\section{Design Criteria}

\begin{figure*}[t]
\setlength{\tabcolsep}{1mm}
\centering
\begin{tabular}{cccc}
\includegraphics[width=0.235\linewidth]{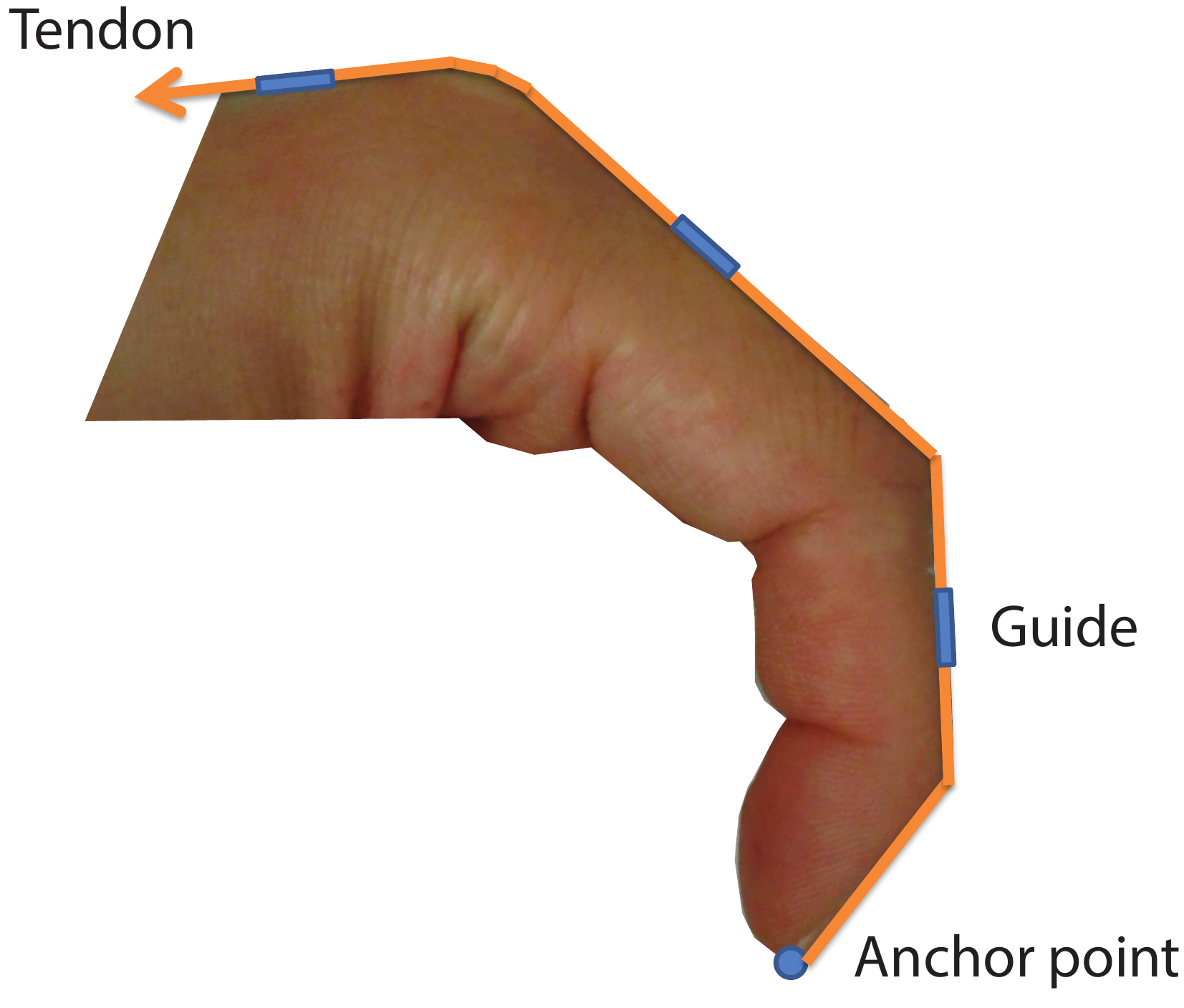}&
\includegraphics[width=0.235\linewidth]{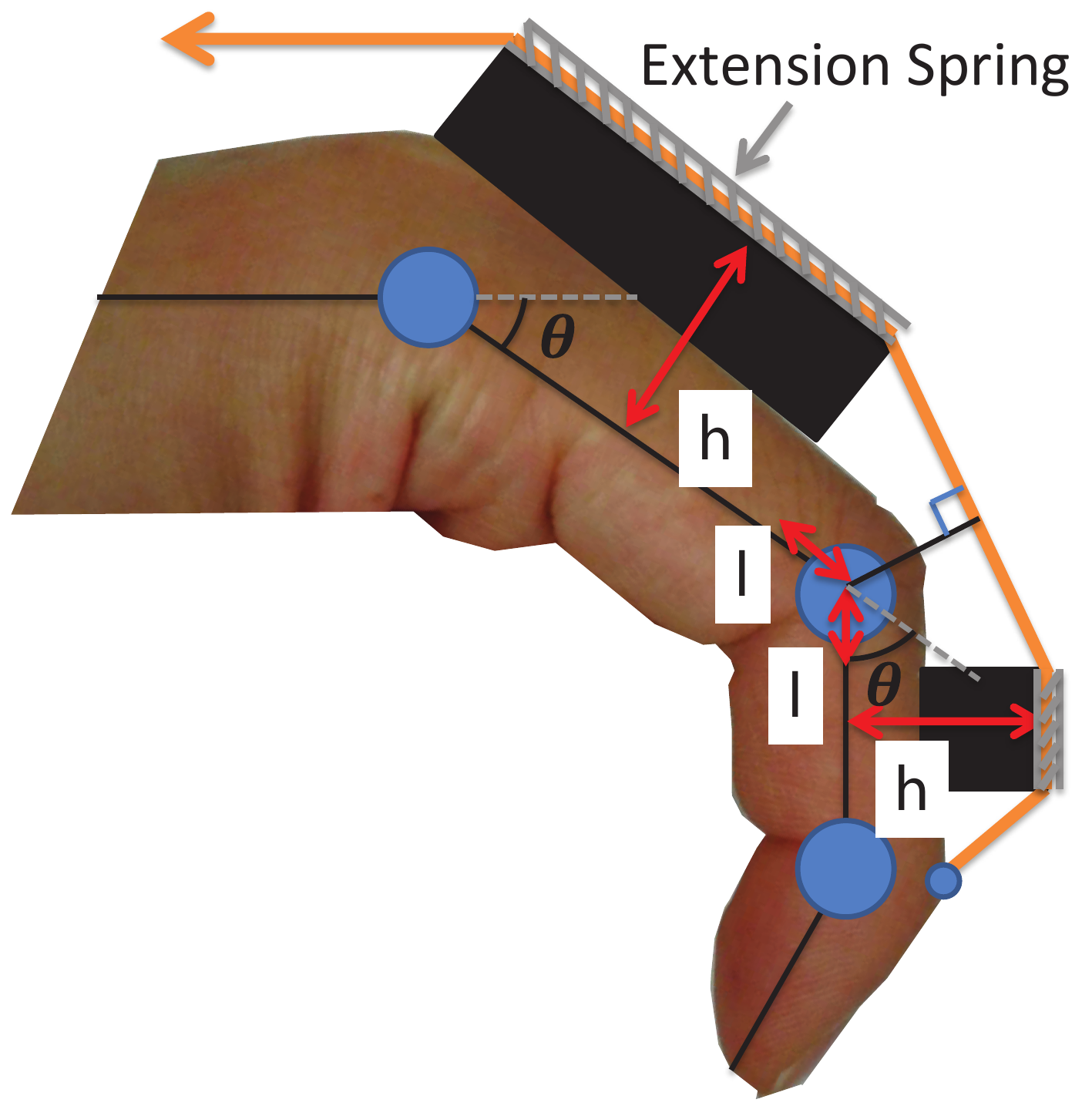}&
\includegraphics[width=0.235\linewidth]{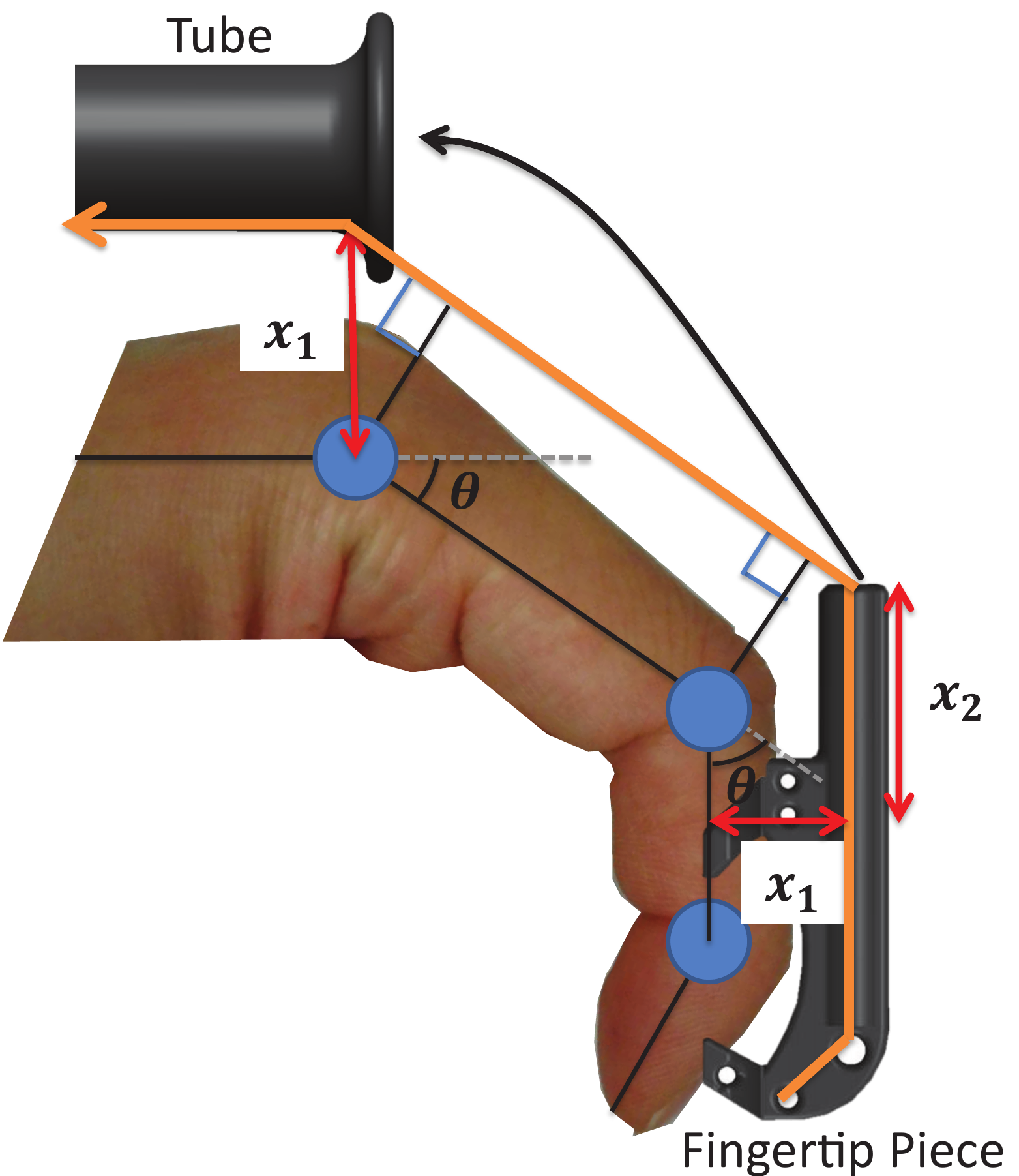}&
\includegraphics[width=0.235\linewidth]{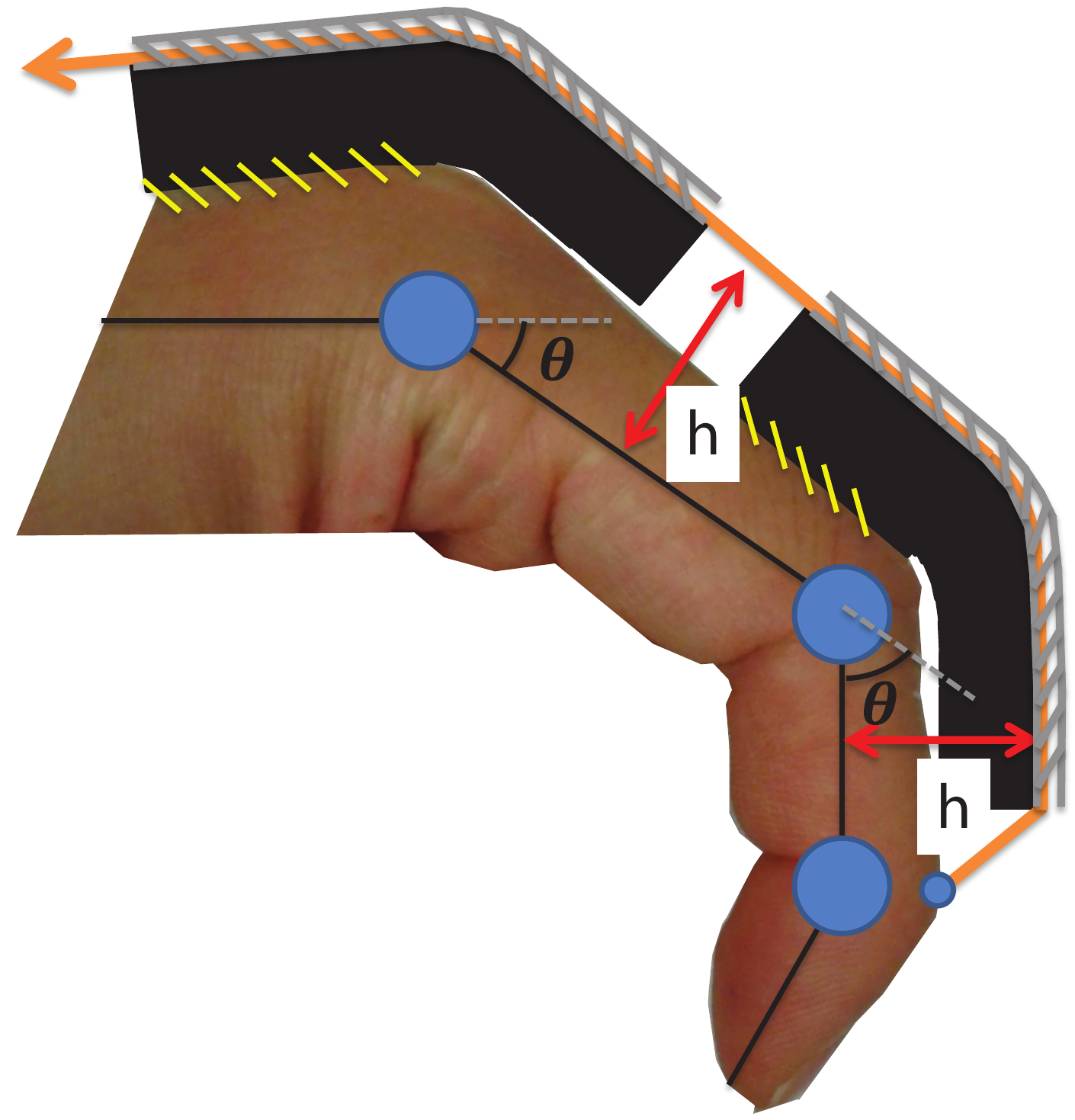}\\
\\[-2mm]
{\footnotesize Traditional Design}&
{\footnotesize Baseline Design}&
{\footnotesize Proposed Design A}&
{\footnotesize Proposed Design B}
\end{tabular}
\caption{\textbf{Traditional Design} : Tubes or rings are installed on the surface of a hand as cable guides, and the fixed point is often located at the finger tip. \textbf{Baseline Design} : Raised pathways are attached on each phalanx, and the fixed point is at the head of middle phalanx. \textbf{Proposed Design A} : A 3D printed part on back of the hand works as an anchor point and another part is attached on the distal and middle phalanx. \textbf{Proposed Design B} : Two raised pathways are used, one between the palm to the proximal phalanx and one between the proximal and middle phalanges. Yellow lines indicate where each pathway segment attaches with the glove. The distal ends of the pathways hang freely, to avoid hindering finger flexion.}
\label{fig:device_illustrations}
\end{figure*}

In this section, we outline the goals that drive our exotendon device development, in a manner independent of specific design choices. In the next section, we present and compare several designs intended to achieve these goals.

\subsection{Achieve Functional Finger Extension}
Impaired finger extension is a common afftereffect in stroke patients. Since finger extension plays an essential role in functional grasp, this impairment adversely affects the quality of life. However, many individuals can still form a grasp in coordinated movement pattern. Given the volitional finger flexion capability, we require that our assistive device help the user achieve functional finger extension.

\subsection{Efficient Transmission}
Given the first requirement above, it follows that exotendons should apply significant extension torques around the MCP and IP joints to overcome spasticity. Use of a strong motor to achieve large extension torques is undesirable as it requires sizable motors and causes distal migration. Increasing tendon moment arms around the joints is an attractive alternative which avoids such unwanted effects. We thus look for effective transmission mechanisms that increase torque for a fixed tendon force.

\subsection{Effective Torque Distribution}
In experiments with stroke survivors Cruz and Kamper~\cite{cruz2010} conducted, proportions of constant extension torques applied to keep the joints in the neutral position were approximately 0.03:0.66:0.46 for the distal interphalangeal (DIP), PIP, and MCP joint respectively. This means that the PIP joint typically exhibits higher tone than the other joints. Therefore, it is important to distribute proper amount of torques translated from a motor force to each joint for some patients.

\subsection{Wearability}
For a device to be used in the home environment, it has to be kept compact and lightweight while delivering meaningful assistance. Designing such hand devices is especially challenging, as available space on a hand is limited. To conform with this constraint, the designs presented here elicit movement using a single motor. In previous work, we have shown that a single motor can elicit the desired movement patterns~\cite{park2016,Meeker2017}, but did not consider the effects of increased spasticity during functional tasks due to repetition and fatigue.

\section{Design and Development} \label{development}


\subsection{Designs}
In this section we introduce a number of possible device designs, which we will
later compare and contrast from the perspective of our
requirements. All the designs discussed here are illustrated in
Fig.~\ref{fig:device_illustrations}.

\begin{figure}[t]
\setlength{\tabcolsep}{1mm}
\centering
\begin{tabular}{cc}
\raisebox{1\height}
{\includegraphics[height=0.45in]{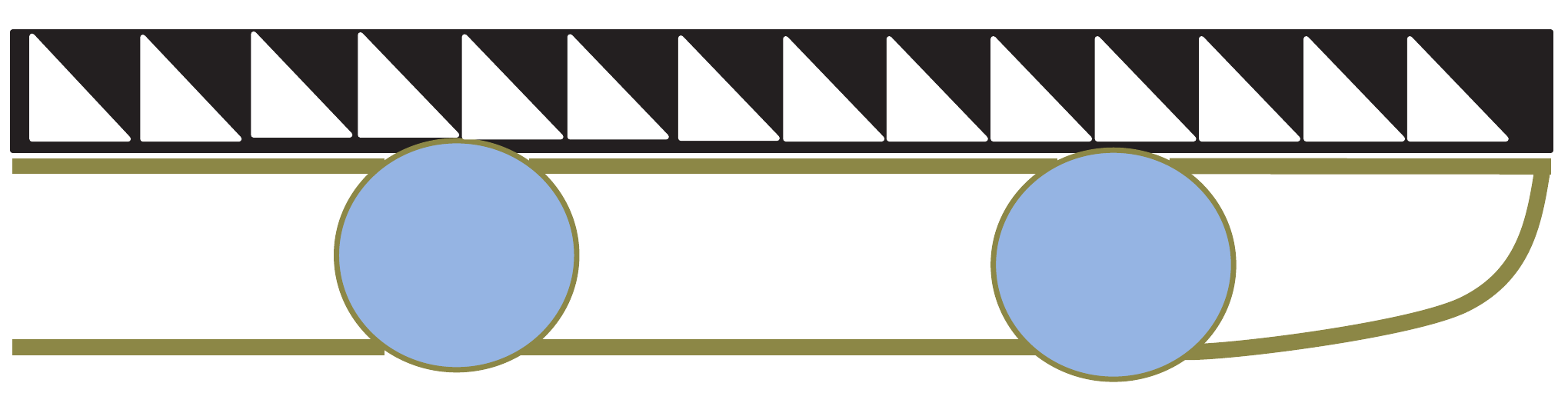}}&
\includegraphics[height=1.5in]{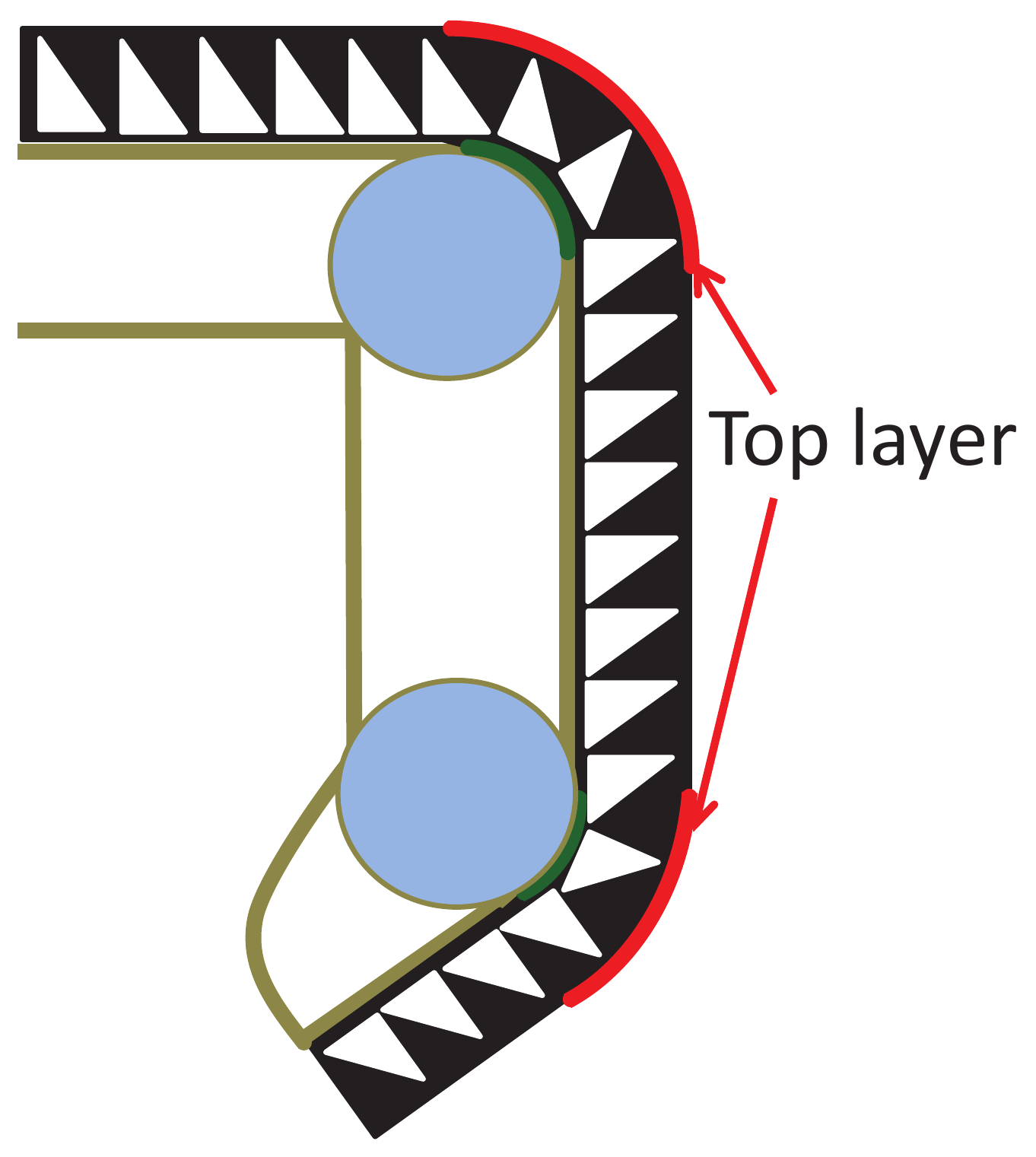}
\end{tabular}
\caption{Example of a raised pathway attached on a finger in extension (left) and flexion (right). Note that insufficient expansion of the top layer of the pathway hinders finger flexion.}
\label{fig:baseline_illustration}
\end{figure}

\begin{figure}[t]
\begin{tabular}{cc}
\hspace*{-2mm}
\raisebox{1\height}
{\includegraphics[height=0.52in]{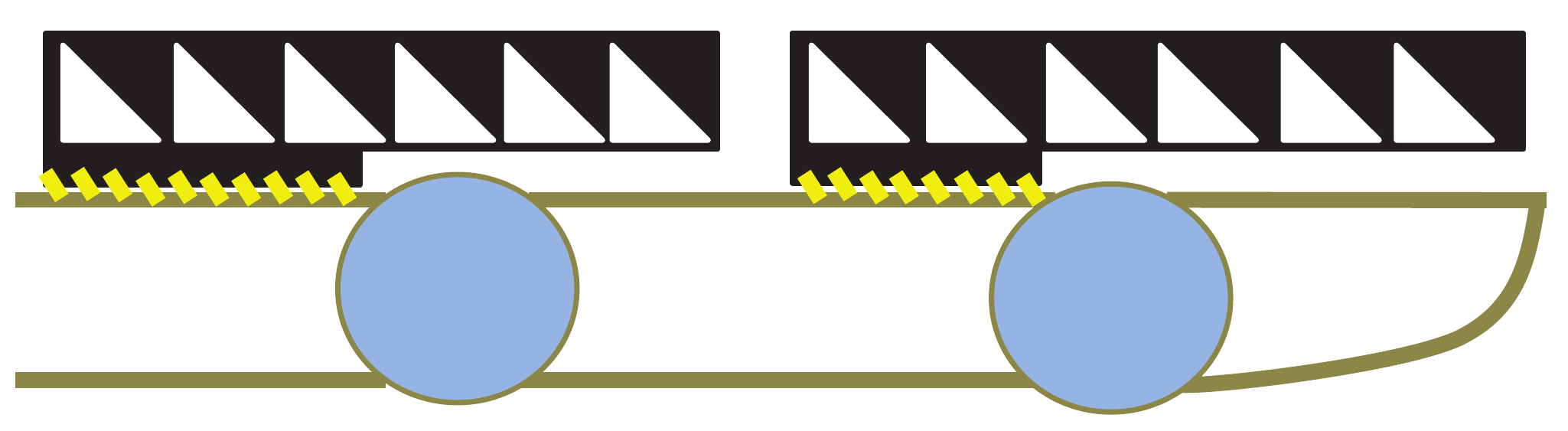}}&
\hspace*{-4mm}
\includegraphics[height=1.5in]{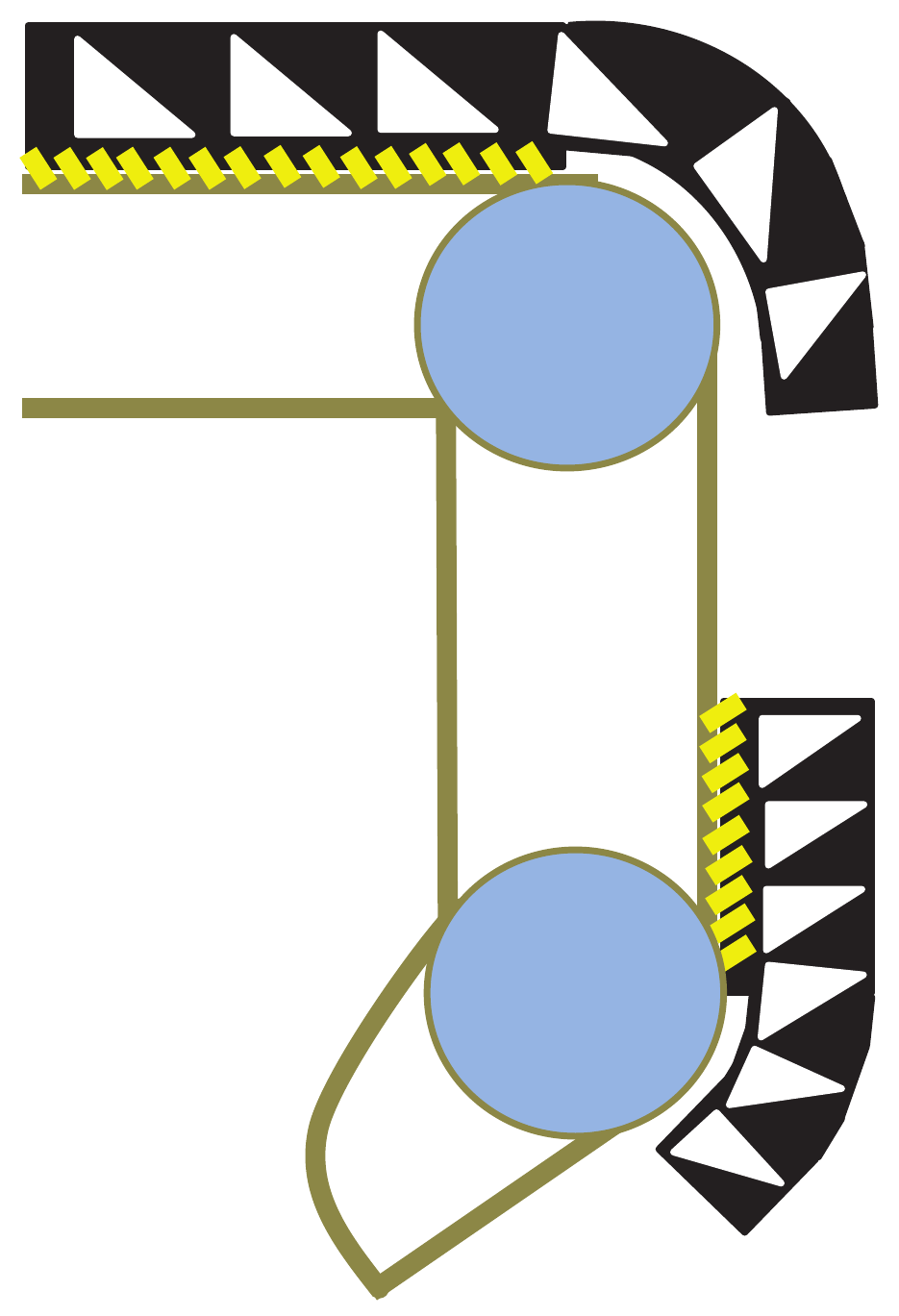}
\end{tabular}
\caption{Simple illustration of Design B on a finger in extension (left) and in flexion (right). Yellow lines show the attachment areas between the glove and the raised pathway. Since the pathway does not elongate in flexion, finger movement is not hindered.}
\label{fig:designB_illustration}
\end{figure}

\subsubsection{Baseline Design}
The starting point is the simplest design where the tendon is simply routed on the dorsal side of the finger (Traditional Design in Fig.~\ref{fig:device_illustrations}). With no moment arm increase, however, this is an ineffective way to achieve the torque levels needed to overcome spasticity and is included here only as a reference starting point. Furthermore, anchoring the tendon at the fingertip is likely to cause hyperextension of a DIP joint unless the range of motion is perfectly fitted with the user. For all other designs, we attach the tendon to the distal end of the middle phalanx.

The most direct way to increase the moment arm is to install a raised pathway for the entire tendon route. However, such a pathway must elongate to support finger flexion, creating elastic effects that hinder motion. This phenomenon increases with the height of the pathway, as the top layer must elongate even further. This behavior is illustrated in Fig.~\ref{fig:baseline_illustration}.

Our Baseline Design thus consists of raised pathways separated section-by-section to avoid the interruption of finger flexion. Although moment arms around the joints are increased with this design, the cable takes a shortcut between the pathways leading to a shorter moment arm when the finger is not fully extended (illustrated in Fig.~\ref{fig:device_illustrations}). In this study, we use this structure as a baseline to make comparisons among envisioned designs. 

\subsubsection{Design A}
The main goal of Design A is to achieve a greater moment arm around the PIP joint, which typically exhibits the strongest spasticity among the three joints of the finger. In this design, we implement two 3D-printed parts as cable guides (Fig.~\ref{fig:device_illustrations}).

A fingertip piece is mainly used to increase a moment arm around the PIP joint. This component also mechanically prevents hyperextension of the DIP joint while assisting finger extension. Fabric straps are secured around the finger using velcro to maintain the position of the device on the hand as depicted in Fig.~\ref{fig:orthosis}. A funnel shaped tube is installed on the dorsal side of the hand to increase the moment arm around the MCP joint. However, for a small handed person, this palmar component may collide with the finger-tip component in full finger extension. Therefore, the funnel tube is designed to allow the fingertip component to be inserted into the dorsal component to enable full range of motion.

Two parameters, ${x_1}$, the normal length between the center of the PIP joint and tendon location and ${x_2}$, the length between a support of the fingertip piece and the end of the fingertip piece, determine the moment arm around the PIP joint depending on the joint angle ${\theta}$.

To learn how the two parameters contribute to the geometric characteristics, we have recorded the moment arm around the PIP joint.
For simplicity, we assume that the PIP and MCP joints are simultaneously moving with the joint angle ${\theta}$, and the range of motion for both joints is from $-90\, ^{\circ}$ (fully flexed) to $0\, ^{\circ}$ (fully extended). Also, we limit the range of ${x_1}$ and ${x_2}$ to avoid designs that are either too bulky or ineffective.

\begin{figure}[t]
\setlength{\tabcolsep}{1mm}
\centering
\begin{tabular}{cc}
\includegraphics[width=0.49\linewidth]{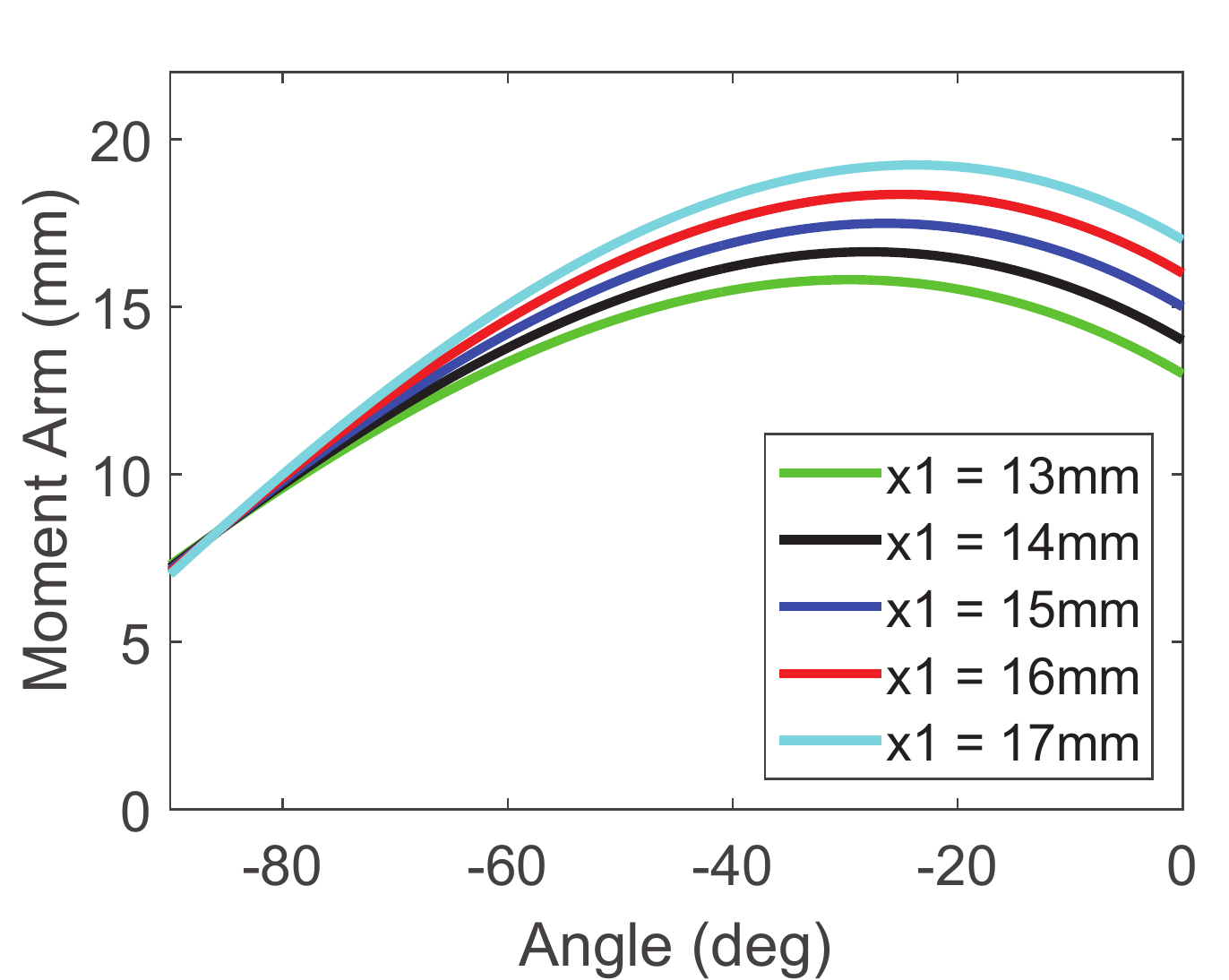}&
\includegraphics[width=0.49\linewidth]{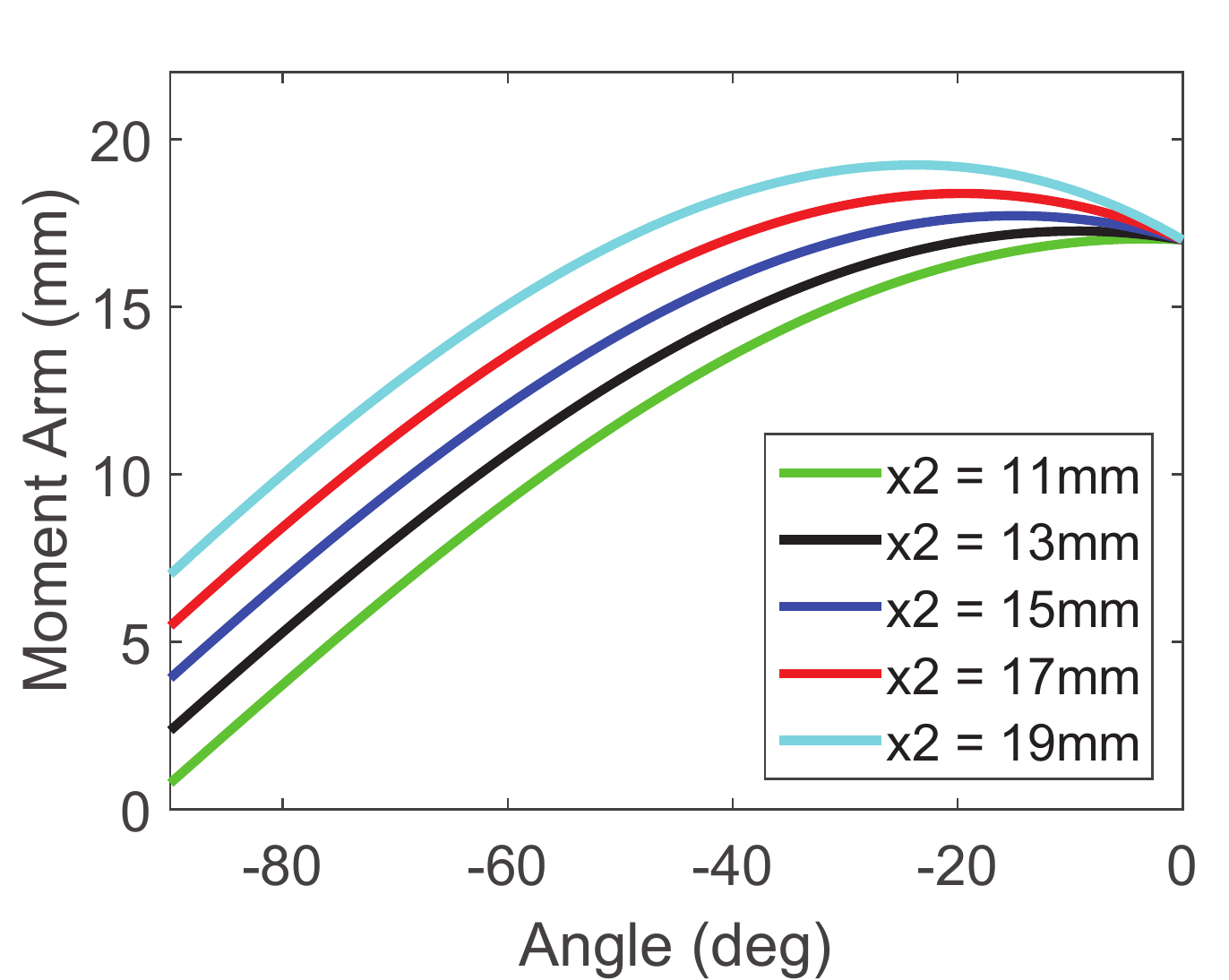}
\\
{\footnotesize (a)}&
{\footnotesize (b)}
\\
\end{tabular}
\caption{(a) Moment arm around the PIP joint vs. joint angle for different ${x_1}$ where ${x_2}$ is fixed at 19mm. (b) Moment arm around the PIP joint vs. joint angle for different ${x_2}$ where ${x_1}$ is fixed at 17mm.}
\label{fig:simulation_DesignA}
\end{figure}

The results indicate that ${x_1}$ is more responsible for torque generation when the finger is extended, whereas ${x_2}$ is more influential for flexed positions, as shown in Fig.~\ref{fig:simulation_DesignA}. With chosen parameters, the PIP and MCP joints are recorded across finger motions with results plotted in Fig.~\ref{fig:simulation_compare}-(a), and it shows that the moment arm around the PIP joint is longer than around the MCP joint.

\subsubsection{Design B}
Design B aims to maintain moment arms around the PIP and MCP joint at preset length ${h}$ and avoid interference with grasps during hand motions. The design consists of a glove, extension springs, and two raised pathways on each finger (Fig.~\ref{fig:device_illustrations}). The pathways are placed on top of the middle and proximal phalanges, and back of the hand covering the PIP and MCP joint. This prevents the tendon from taking a shortcut that reduces the moment arms around the two joints. In order to avoid hindering finger flexion, raised pathways are rigidly secured proximally to the PIP and MCP joint while the other ends distally located to the joint are free to slide without direct attachment(Fig.~\ref{fig:designB_illustration}). A cloth cover is sewn on top of the pathway to prevent it from drifting laterally during use.

\subsection{Design Comparison}
\begin{figure}[t]
\setlength{\tabcolsep}{1mm}
\centering
\begin{tabular}{cc}
\includegraphics[width=0.5\linewidth]{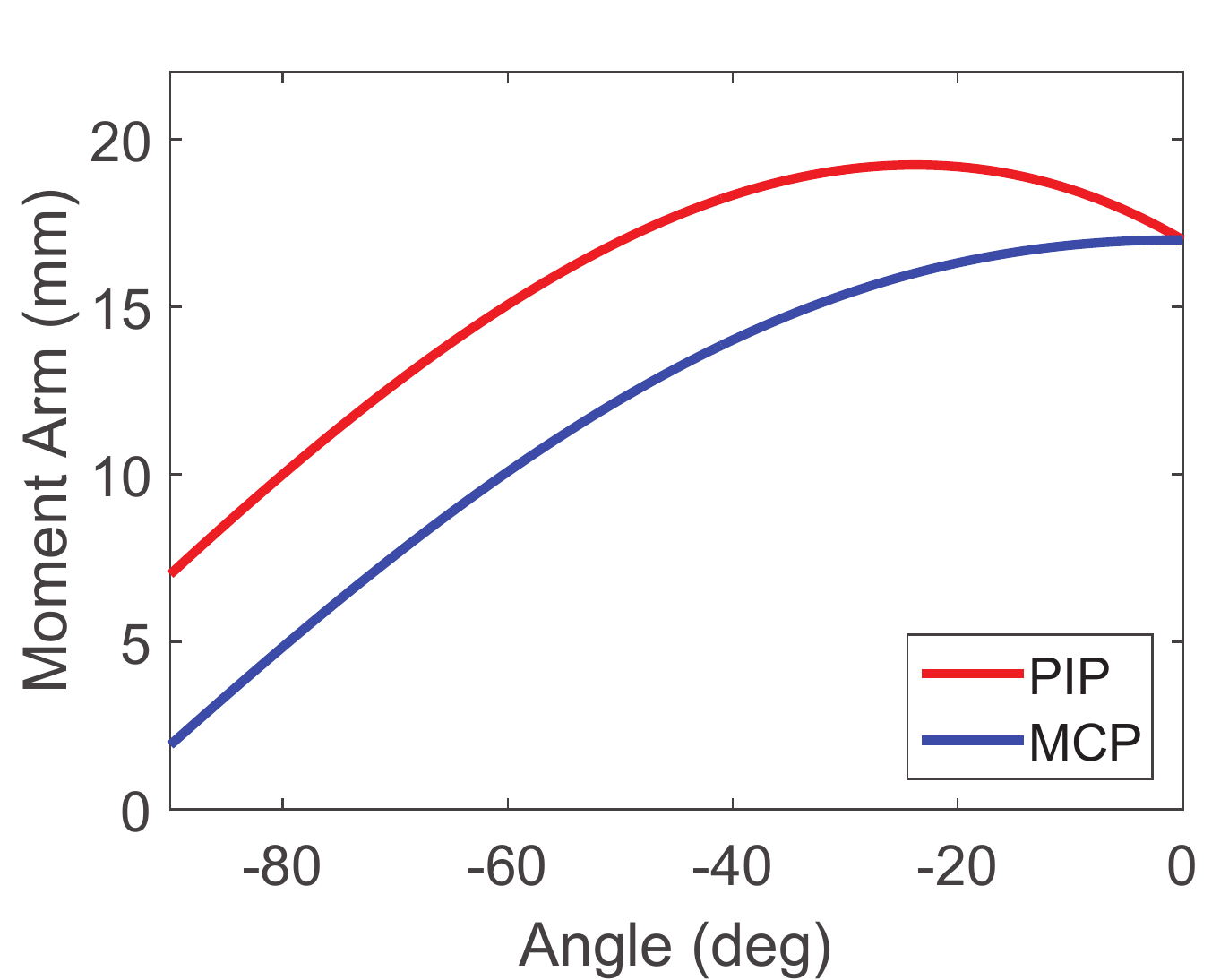}&
\includegraphics[width=0.5\linewidth]{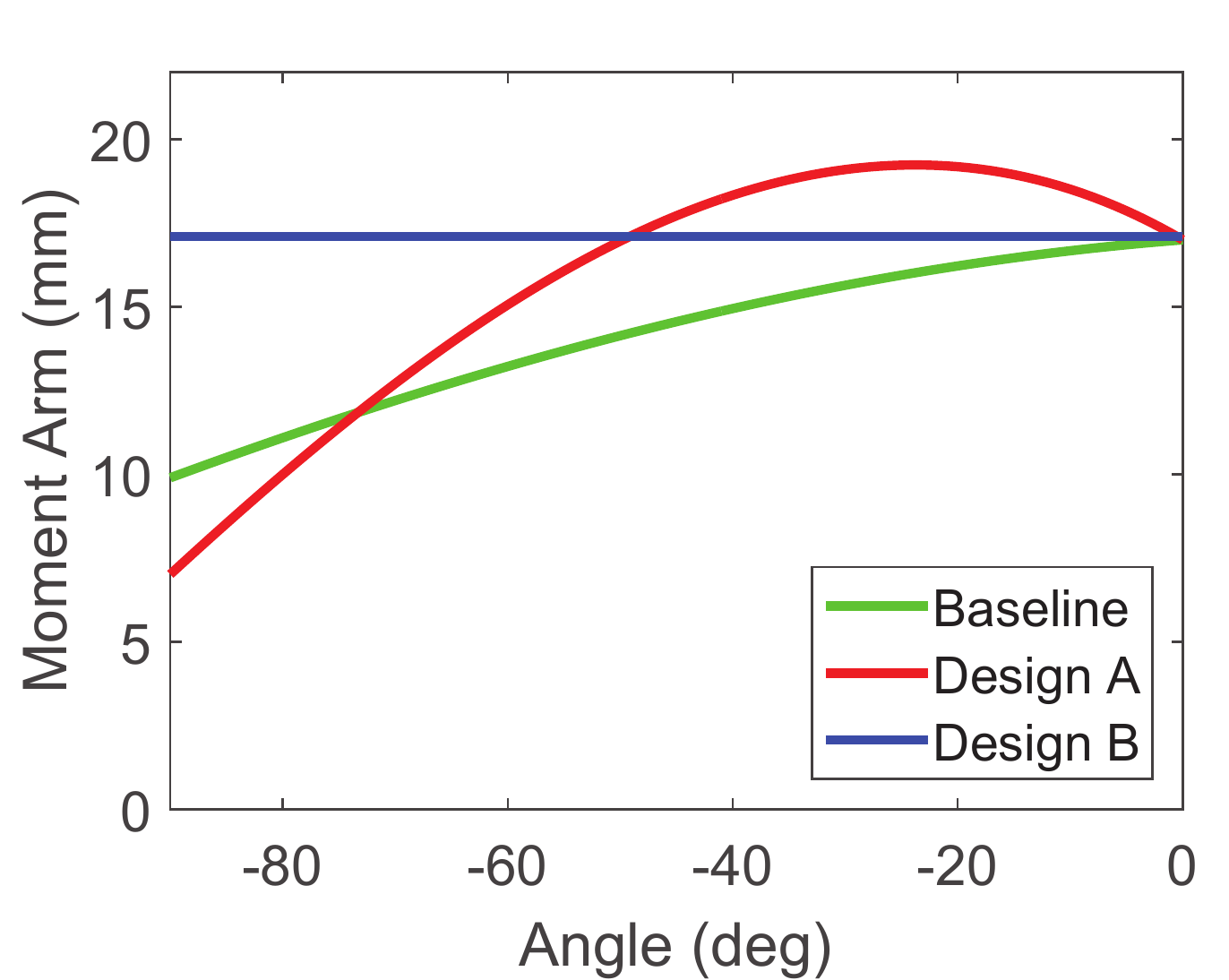}
\\
{\footnotesize (a)}&
{\footnotesize (b)}
\\
\end{tabular}
\caption{(a) Moment arm around the PIP and MCP joint vs. joint angle for Design A. (b) Moment arm around the PIP joint vs. joint angle for Baseline design, Design A, and Design B. }
\label{fig:simulation_compare}
\end{figure}

Fig.~\ref{fig:simulation_compare}-(b) shows a moment arm around the PIP joint versus joint angle ${\theta}$ for Baseline design, Design A, and Design B. The length between the center of each joint and the tendon is set to 17mm at fully extended position for all three designs.

The result indicates that Design A and Design B generate a larger moment arm around the PIP joint than Baseline design throughout hand motions. The moment arm with Design A in an extended finger position is notably larger than the others, which is beneficial considering that a proportional increase in extension torque is required as the finger is in a more extended position~\cite{park2016}. Design B also creates a larger moment arm than Baseline design.

From the outcomes, one can assume that the force level required to extend the finger by Baseline design would be the greatest. For Design A, the extension would require relatively lower force level on the PIP joint than the MCP joint as the moment arm around the PIP joint is larger. Finally, in Design B, the PIP and MCP joints would need similar level of force to execute full extension because the geometry of the design for both joints is relatively similar.

\subsection{Exotendon Device}
The designs described above are used in a combination with the exotendon device previously developed in our lab. Mechanical components of the device are composed of a
forearm piece with actuation and a structure based on the two designs (Fig.~\ref{fig:orthosis}). This structure engages the impaired hand with a motor on the forearm piece through a tendon network. An S-hook connects the tendon network from the end effector with the motor to facilitate the donning process.

The forearm piece works as both an anchor point to stabilize a base of the motor and a splint that constrains the wrist movement to efficiently transmit the motor force to the end effector. A DC motor (Pololu corporation, 47:1 Medium-Power 25D Metal Gearmotor) with a 100N peak force is mounted on the forearm piece. The motor is driven by Proportional-Integral-Derivative(PID) position controller, and the range of motion is determined at the clinical test after fitting the device. A simple push button is implemented to trigger finger extension. While pushing the button, the motor stalls when the applied motor force reaches its maximum level or the fingers arrive at the fully extended position. Releasing the button allows the fingers to flex and the hand to grasp.

The DC motor applies an extension force to all four fingers except for the thumb, which moves in unique ways compared with the other fingers. For grasping tasks, splinting the thumb in a functional, opposed position is sufficient if the four fingers are adequately extended by the device~\cite{arata2013}. The specific thumb splinting and device controlling approaches, other than the aforementioned button control, are excluded from this study as they are out of scope. We are planning to report these methods in the near future.

\section{Experiments and Results}
\label{experiments}
In order to evaluate the theoretical results shown in the previous section, we have conducted experiments using a 3D-printed artificial finger to find a relation between joint angles and applied force. In addition, we performed clinical tests with stroke patients to provide validity of the results from simulations and experiments with an artificial finger. In the experiments, subjects wore a device with each of Baseline design, Design A, and Design B at a time, and the joint angles of an index finger were measured while the device was assisting finger extension. We provide a comparison of the range of finger extension elicited by the three designs.

\subsection{Testing with Artificial Finger}
\begin{figure}[t]
\setlength{\tabcolsep}{1mm}
\centering
\begin{tabular}{c}
\includegraphics[width=1\linewidth]{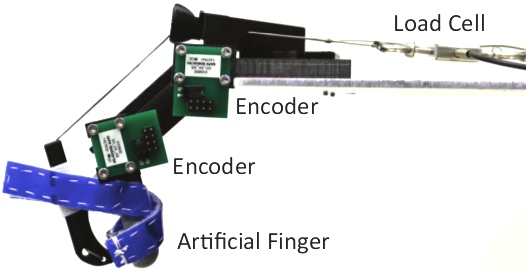}
\\
\end{tabular}
\caption{Experimental set-up with an artificial finger.}
\label{fig:artificial_finger}
\end{figure}

The artificial finger consists of 3D printed parts, torsion springs, and encoders(Fig.~\ref{fig:artificial_finger}). Torsion springs are installed on all three joints to mimic hand spasticity. A proportion of the spring constants is 3.5:76.9:54.9 for the DIP, PIP, and, MCP joint respectively, which is similar to 0.03:0.66:0.46 from Cruz and Kamper's work~\cite{cruz2010}. Hall effect rotary encoders(AS5600) are placed on the side of the PIP and MCP joint. The DIP joint, which is coupled with the PIP joint is excluded from the measurements for the sake of simplicity. Since four fingers other than the thumb have similar structures and exhibit identical movements experiments with one fingered device should suffice.

The main objective of this experiment is to measure force characteristics throughout an entire hand motion assisted by the device with Baseline design, Design A, and Design B. To measure a tension on the actuated tendon, a load cell(Futek, FSH00097) is installed in series with a motor and the tendon network. The force is recorded at 100Hz while the motor applies extension torques to the fully flexed artificial finger ($-90\, ^{\circ}$) till the finger is fully extended ($0\, ^{\circ}$).

\begin{figure}[t]
\setlength{\tabcolsep}{1mm}
\centering
\begin{tabular}{c}
\includegraphics[width=1\linewidth]{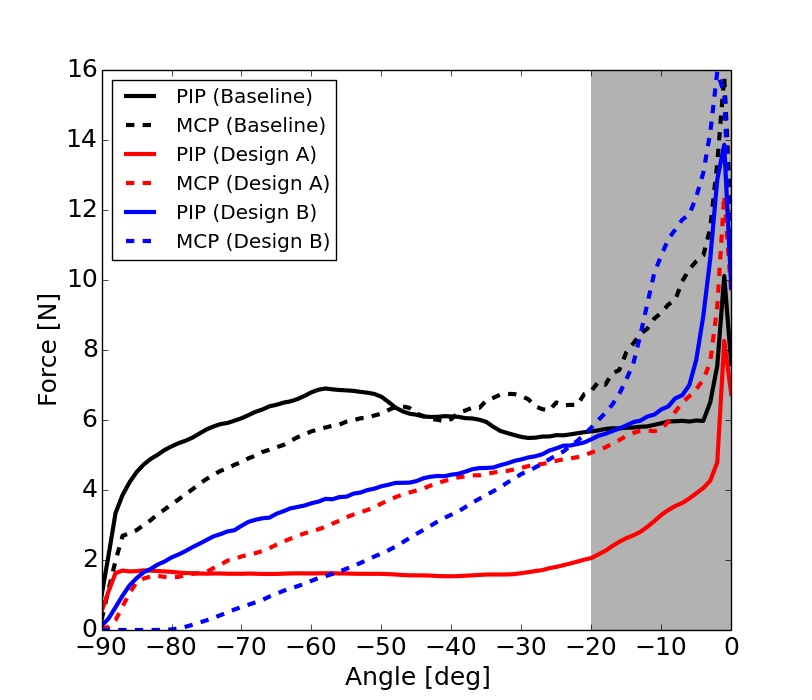}
\\
\end{tabular}
\caption{Force vs. PIP (solid) and MCP (dashed) joint angles from artificial finger experiments. Black, blue, and red colors indicate force needed to reach certain joint angles with Baseline design, Design A, and Design B respectively. Particular ranges of our interest span from $-90\, ^{\circ}$ to $-20\, ^{\circ}$ as this region encapsulates the necessary functional movements.}
\label{fig:artificial_finger_plot}
\end{figure}

For reliability, the measurements have been taken 50 times for each design. The average values of force vs. joint angles are shown in Fig.~\ref{fig:artificial_finger_plot}. The result shows higher force is required for Baseline design than the others to achieve finger extension. Also, note that differences in force requirement are more prominent in flexed positions as simulation results from the previous section suggest. For Design A, the PIP joint requires less force than the MCP joint across hand motions as evidenced from Fig.~\ref{fig:simulation_compare}-(a).

\begin{figure}[t]
\setlength{\tabcolsep}{1mm}
\centering
\begin{tabular}{c}
\includegraphics[width=1\linewidth]{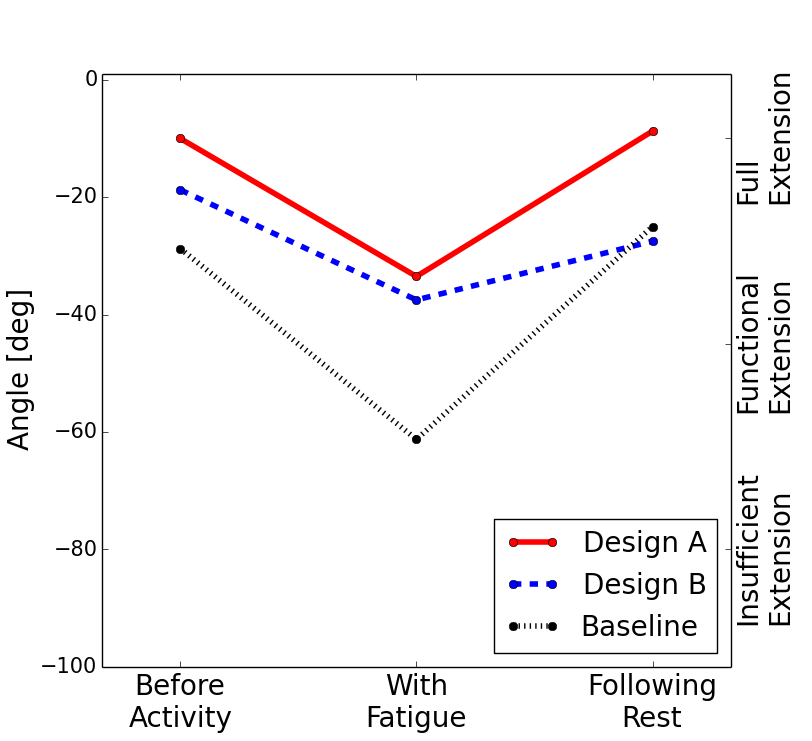}
\\
\end{tabular}
\caption{Joint angles of the index finger measured with stroke patients while the device with Baseline design(black dotted line), design A(red solid line), and design B(blue dashed line) is assisting finger extension.}
\label{fig:clinical_test_result}
\end{figure}

\begin{table}[]
\vspace{2mm}
\centering
\caption{Subject clinical information}
\label{MAS_table}
\begin{tabular}{C{0.7cm}|C{.6cm} C{.6cm} C{.6cm}|C{.6cm} C{.6cm} C{.6cm}}
        	& \multicolumn{3}{c|}{MAS Extensor Score}     	& \multicolumn{3}{c}{MAS Flexor Score}    \\ 
Subject & Elbow  	& Wrist 	& Finger	& Elbow		& Wrist		& Finger   \\ \hline
A       & 1+         	& 0      	& 0         	& 1		& 0		& 1           \\ 
B       & 2         	& 0      	& 0          	& 2		& 1		& 1+            \\ 
C       & 1          	& 0        	& 1           	& 1+		& 1		& 1+            \\ 
D       & 0          	& 0        	& 0           	& 0		& 0		& 1            \\
\end{tabular}
\end{table}

\subsection{Testing with Stroke Patients}
Four participants, one female and three male, with right side hemiparesis and limited mobility following a stroke event at least 6 months prior were recruited from a voluntary research registry of individuals who have survived stroke. Testing was approved by the Columbia University Institutional Review Board and took place in a clinical setting under the supervision of licensed physical and/or occupational therapists. Upper limb spasticity measurements were between 0 and 2 on the Modified Ashworth Scale(MAS) for all participants (Table~\ref{MAS_table}).

Each testing session was performed over the course of one visit. Spasticity scores at the elbow, wrist, and digits were assessed using the MAS before and after testing. Subjects were then fitted with the exotendon orthotic device and guided through the following procedure with each version of the device.

\begin{itemize}
\item The subject opens the hand using the orthosis. Extension of the PIP and MCP joints of the index finger are measured using a goniometer. The index finger was selected for measurement as it was most accessible with the device in place; it was also qualitatively observed to be representative of the four fingers.
\item The subject attempts to grasp and release 15 times to induce fatigue. The device is triggered to assist hand opening using a button at the point of maximal effort. On the 15th repetition, the joint angles of the index finger are measured again. In general, fatigue increases tone in hand movement, and this measurement is taken to see if one can still achieve functional hand extension in this condition.
\item The subject takes a rest for five minutes to reduce the impact of fatigue. Then, the last measurements of the joint angles of the index finger are recorded while the device is assisting. 
\item To avoid effects of fatigue carrying over to the next trial, the subject rests for ten minutes between trials with different devices (including time spent on doffing and donning the devices).
\end{itemize}

Fig.~\ref{fig:clinical_test_result} shows the average of measured PIP joint angles with all participants. Since the MCP joint was fully extended for every patient, only the PIP joint angle was measured.

The result demonstrates a comparative advantage of Design A and Design B over Baseline design. In particular, as patients became fatigued, the Baseline design generally failed to elicit functional extension, whereas Design A and Design B were less vulnerable to increased tone after activities. This result also suggests that assessing the feasibility of a hand device through range of motion measurements without the integration of functional tasks may not be representative of real life use. For a hand device to allow repetitive exercises, post-fatigue evaluation should also confirm the effectiveness of the device.

\section{Conclusions}

\begin{table}[]
\caption{Joint angles of the index finger measured with stroke patients(mean$\pm$ standard error)}
\label{clinical_test_table}
\centering 
\begin{tabular}{c | c c c}
Version & Before Activity & With Fatigue & Following Rest\\ [0.5ex]
\hline
\\[-2.75mm]
\hline
\\[-2mm]
Design A & -10.0($\pm$5.7) & -33.5($\pm$6.4) & -8.8($\pm$5.9) \\
Design B & -18.8($\pm$11.9) & -37.5($\pm$20.5) & -27.5($\pm$16.0)  \\
Baseline Design & -28.8($\pm$12.6) & -61.3($\pm$18.1)  & -25.0($\pm$16.0)  
\end{tabular}
\label{table:nonlin}
\end{table}

In this study, we have proposed distal structures of a tendon driven hand assistive orthosis for efficient force transmission. In order to evaluate the designs, we ran simulations with mathematical models and conducted experiments using a 3D-printed artificial finger. We also performed clinical tests with stroke patients to study real-life applicability.

In the geometric model analysis, the two designs we propose enabled torque generations with large moment arms around joints of a finger, compared to traditional devices. The advantages of the large moment arm were demonstrated by the experiments with an artificial finger. The result suggests that the force required by proposed structures to extend a finger was lower than Baseline design. Clinical trials with four stroke patients where we measured joint angles with the device assisting finger extension also supported the feasibility of the effective mechanism. In the experiments, all four participants attained a functional finger range of motion even when fatigued with the assistance of Design A and Design B.

Efficient transmission mechanisms in a wearable tendon driven device offer several advantages. Since the required level of force is lower, the size of the device can be reduced, and the likelihood of injury decreased. Small and light actuators can be placed closer to the affected hand, further increasing wearability. In addition, a low tendon force leads to reduced distal migration, which is a limitation of active devices.

A limitation of the assessment with patients with stroke is potential inaccuracy from the use of a manual tool such as a goniometer. Future work will also involve more rigorous evaluations in functional performance, since increased range of motion may not guarantee improved performance of grasp/release tasks or activities of daily living. 

Another potential limitation is the ability to adjust the fit of each design to the dimensions of the palm and digits of the patient. Throughout our experiments, we learned that this fit plays an important role in performance. Accordingly, future prototypes will include additional adjustable components; this is especially important for Design A because it includes rigid parts which make fitting more difficult. Once the fit is stabilized, we also plan to conduct experiments with stroke patients to demonstrate functional improvements with the device in integration with more intuitive control methods, such as forearm electromyography. We believe that the combination of efficient mechanical designs with intuitive control methods can lead to wearable devices used for numerous activities, providing functional assistance and improving quality of life.

\bibliographystyle{IEEEtran}
\bibliography{bib/orthoses,bib/stroke}
\end{document}